%% file: main.tex
\documentclass[letterpaper]{article}
\usepackage{aaai}
\usepackage{amsmath}
\usepackage{amssymb}
\usepackage{times}
\usepackage{helvet}
\usepackage{courier}
\usepackage{natbib}
\usepackage{hyperref}
\usepackage{listings}
\usepackage{subcaption}
\usepackage{environ}
\hypersetup{
     colorlinks   = true,
     linkcolor=blue,
     citecolor    = blue 
}

\usepackage[english, status=draft]{fixme}
\fxusetheme{color}

\usepackage[acronym]{glossaries}
\newacronym[longplural={Gaussian Processes}]{gp}{GP}{Gaussian Process}
\newacronym{ckl}{CKL}{Compositional Kernel Learning}
\newacronym{abcd}{ABCD}{Automatic Bayesian Covariance Discovery}
\newcommand{\gp}{\mathcal{GP}}
\usepackage{tikz}
\usetikzlibrary{bayesnet, calc,positioning,shapes.geometric}
\makeatletter
\newsavebox{\measure@tikzpicture} 
\NewEnviron{scaletikzpicturetowidth}[1]{%
  \def\tikz@width{#1}%
  \begin{lrbox}{\measure@tikzpicture}%
  \BODY
  \end{lrbox}%
  \pgfmathparse{#1/\wd\measure@tikzpicture}%
  \BODY
}
\makeatother

\begin{document}
\title{Automatic Generation of Probabilistic Programming from Time Series Data}
\author{Anh Tong \and Jaesik Choi\\
Ulsan National Institute of Science and Technology\\
Ulsan, 44919 Korea\\
\{anhth,jaesik\}@unist.ac.kr}
\nocopyright
\maketitle

\begin{abstract}
\begin{quote}
Probabilistic programming languages represent complex data with intermingled models in a few lines of code. Efficient inference algorithms in probabilistic programming languages make possible to build unified frameworks to compute interesting probabilities of various large, real-world problems. When the structure of model is given, constructing a probabilistic program is rather straightforward. Thus, main focus have been to learn the best model parameters and compute marginal probabilities. In this paper, we provide a new perspective to build expressive probabilistic program from continue time series data when the structure of model is not given.
The intuition behind of our method is to find a  descriptive covariance structure of time series data in nonparametric Gaussian process regression. We report that such descriptive covariance structure efficiently derives a probabilistic programming description accurately.
\end{quote}
\end{abstract}

\section{Introduction}
\input{./1-intro.tex}

\section{Background}
\input{./2-background.tex}

\section{Automatic Generation of Probabilistic Programming}
\input{./3-combi.tex}

\section{Experiment}

\input{./4-experiment.tex}

\section{Conclusion}
\input{./5-conclusion.tex}

\section*{Acknowledgments} 
This work is supported by Basic
Science Research Program through the National Research
Foundation of Korea (NRF) grant funded by the Ministry
of Science, ICT \& Future Planning (MSIP) (NRF-
2014R1A1A1002662) and the NRF grant funded by the MSIP
(NRF-2014M2A8A2074096).
\bibliographystyle{aaai}
\bibliography{ref.bib}

\input{./6-appendix.tex}
\pagebreak
\end{document}

%% file: 1-intro.tex
Probabilistic programming has potential impacts on various works in artificial intelligence, machine learning, statistics, robotics (\cite{robotLBDM04}), vision (\cite{Kulkarni_2015_CVPR}), neuroscience (\cite{neuroLake1332}), and cognitive science (\cite{cogniFRT12}).
Many different approaches and probabilistic programming languages are introduced, for example, Church \citep{church}, Problog \citep{problog}, BUGS \citep{bugs},  \texttt{Stan} \citep{stan}. Although each probabilistic programming language has its own strength and domain, we choose \texttt{Stan} to present our work in this paper since it is suitable for modeling continuous signal.

The Automatic Bayesian Covariance Discovery (ABCD) system which is so-called automatic statistician system, is proposed \citep{lloyd2014automatic} with the aim of automating the process of statistical modeling. It focuses on regression problems. Specifically, it takes time series as input, searches and then produces a learned Gaussian process model which has interpretable properties (smoothness, periodicity, changepoints) and is summarized with a report. With the same purpose, \citep{1511.08343} proposed a relational approach to handle multiple data. However, such kinds of modeling usually require a body of work and efforts to make build a new model. One of the advantages of probabilistic programming which helps in this situation is the ease of creating generative models with several lines of code \citep{Kulkarni_2015_CVPR}. In order to facilitate the process of building new ABCD-based models, we propose a method generating \texttt{Stan} probabilistic programming from ABCD results. Time series are stored in the compact representation of encoded ABCD probabilistic programmings which potentially allow construct a complex model from heterogeneous models.

This paper is organized as follows. We briefly introduce the background of Gaussian Processes (GP) , the ABCD system and its relational version for multiple data. Then, we present our main contribution on how a probabilistic programming is generated from time series data; Next is the experiment section; Finally, we conclude our work.

%% file: 2-background.tex
\subsection{Automatic Statistician System}
\input{./2.1-abcd.tex}
\subsection{Relational Automatic Statistic System}
\input{./2.2-rabcd.tex}
\subsection{Probabilistic Programming}

\input{./2.3-probabilistic-programming.tex}

%% file: 2.1-abcd.tex
Automatic Bayesian Covariance Discovery (ABCD) explains data without requiring expert input for regression problems. Here the approach is to use Gaussian processes to model regression functions.

Let us take a brief overview about Gaussian Processes (GPs). GPs are distributions over functions such that any finite set of function evaluations, $f(x) = (f(x_1), f(x_2), \dotsc, f(x_N))$ form a multivariate Gaussian distribution \citep{rasmussen2006gaussian}. It is specified by a mean function $\mu(x) = \mathbb{E}[f(x)]$ and a covariance kernel function $k(x, x') = \mathrm{Cov}$(f(x), f(x')). Evaluations of the two functions on a finite set of points correspond to the mean vector and the covariance matrix for the multivariate Gaussian distribution, like $\mathbf{\mu}_i{=}\mu(x_i)$ and $\mathbf{\Sigma}_{ij}{=}k(x_i, x_j)$. When a function or evaluations of the function $f$ are drawn from a \gls*{gp} specified by its mean function $\mu(x)$ and covariance kernel function $k(x,x')$, 
\begin{align*}
f \sim \gp{\left(\mu(x),k(x,x')\right)}
\end{align*}
Covariance kernel function plays a crucial role in a GP, as it conveys our assumptions about the function. \cite{duvenaud4922structure} proposed a compositional kernel learning method. It constructs and find richer kernel which are composed of server base kernels and operation. In theory, any positive definite kernels are closed under addition and multiplication.

Five base kernels are used for making compositional kernels. Each kernel encodes different characteristics of functions, which further enables the generalization of structure and inference given new data.
\begin{tabular}{c | c}
Base Kernels & Encoding Function \\
\hline
White Noise (\textsc{WN}) & Uncorrelated noise \\
Constant (\textsc{C}) & Constant functions \\
Linear (\textsc{LIN}) & Linear functions \\
Squared Exponential (\textsc{SE}) & Smooth functions \\
Periodic (\textsc{PER}) & Periodic functions
\end{tabular}

The first operation is addition which sums multiple kernel functions and makes a new kernel function.
\begin{align*}
k'(x,x') = k_1(x,x') + k_2(x,x').
\end{align*}
ABCD assumes zero mean for \glspl*{gp}, since marginalizing over an unknown mean function can be equivalently expressed as a zero-mean \gls*{gp} with a new kernel \citep{lloyd2014automatic}. Under this assumption, the following multiplication operation is also applicable.
\begin{align*}
k'(x,x') = k_1(x,x') \times k_2(x,x').
\end{align*}
The third operation is the change-point (CP) operation. Given two kernel functions, $k_1$ and $k_2$, the new kernel function is represented as follows:
\begin{align*}
\begin{split}
&k'(x,x') = \sigma(x)k_1(x,x')\sigma(x') \\
&\quad+ (1-\sigma(x))k_2(x,x')(1-\sigma(x')) \nonumber
\end{split}
\end{align*}
where $\sigma(x)$ is a sigmoidal function which lies between 0 and 1, and $\ell$ is the change-point.
The change-point operation divides function domain (i.e., time) into two sides and applies different kernel function on each side. 

Finally, the change-window (CW) operation applies the CP operation twice with two different change points $\ell_1$ and $\ell_2$. Given two sigmoidal functions $\sigma_1(x; \ell_1)$ and $\sigma_2(x; \ell_2)$ where $\ell_1 < \ell_2$, the new function will be $f:=\sigma_1(x)f_1(1-\sigma_2(x)) + (1-\sigma_1(x))f_2\sigma_2(x)$, which applies the function $f_1$ to the window $(\ell_1, \ell_2)$. 
A composite kernel expression after the change-window operation will be as follows:
\begin{align*}
&k'(x,x') = \sigma_1(x)(1-\sigma_2(x))k_1(x,x')\sigma_1(x')(1-\sigma_2(x')) \nonumber \\
&\quad+ (1-\sigma_1(x))\sigma_2(x)k_2(x,x')(1-\sigma_1(x'))\sigma_2(x'). \nonumber
\end{align*}

ABCD searches a composite kernel based on the search grammar. The search grammar specifies how to develop the current kernel expression by applying the operations with the base kernels. The following rules are examples of typical search grammar:
\begin{align*}
\mathcal{S} &\rightarrow \mathcal{S + B} &&\mathcal{S} \rightarrow \mathcal{S \times B} \\
\mathcal{S} &\rightarrow \mathrm{CP}(\mathcal{S,S}) &&\mathcal{S} \rightarrow \mathrm{CW}(\mathcal{S,S})  \\
\mathcal{S} &\rightarrow \mathcal{B} &&\mathcal{S} \rightarrow \mathrm{C} 
\end{align*}
where $\mathcal{S}$ represents any kernel subexpression, $\mathcal{B}$ and $\mathcal{B}'$ are base kernels. 

Given data and a maximum search depth, the algorithm gives a compositional kernel $k(x,x';\theta)$. Starting from the \textsc{WN} kernel, the algorithm expands the kernel expression based on the search grammar, optimizes hyperparameters for the expanded kernels, evaluates those kernels given the data and selects the best one among them. This procedure repeats. The next iterative procedure starts with the best composite kernel selected in the previous iteration. The conjugate gradient method is used when optimizing hyperparameters. Bayesian Information Criterion (BIC) is used for the model evaluation. The BIC of model $\mathcal{M}$ with $|\mathcal{M}|$ number of free parameters and data $\mathcal{D}$ with $|\mathcal{D}|$ number of data points is:
\begin{equation}
\label{eq:bic}
\textsc{BIC}(\mathcal{M}) = -2\log p(\mathcal{D}|\mathcal{M}) + |\mathcal{M}| \log |\mathcal{D}|.
\end{equation}
The iteration continues until the specified maximum search depth is reached. During the iteration, the algorithm keeps the best model for the output.

%% file: 2.2-rabcd.tex
\cite{1511.08343} has developed a relational version to deal with multiple sequences of data. The assumption is that sequences are related to each other. Relational Automatic Bayesian Covariance Discovery (Relational ABCD) finds a shared structure across multiple sequences.

\begin{figure}
\centering
\begin{tikzpicture}

  \node[obs]                               	(y) {$y_i$};
  \node[latent, left=0.5cm of y] 				(fi) {$f_i$};
  \node[latent, left= 0.5cm of fi]  			(f) {$f$};
  \node[const, left=0.5cm of f]            	(gp) {$\mathcal{GP}$};
  \node[latent, left=0.7cm of gp]				(s) {$\sigma_j$};
  \node[const, below=0.5cm of gp]				(kj) {$k_{d_j}(x,x')$};
  \node[const, above=0.6cm of gp]				(ks) {$k_S(x,x')$};
  \node[const, left=0.5cm of ks]				(mu) {$\mu(x)$};
  \node[const, above=0.6cm of mu]				(zero) {$0$};
  \node[latent, right=0.4cm of zero]				(k_s) {$k_S$};
  \node[latent, right =1.6cm of zero]				(theta) {$\theta$};
   \node[const, above=0.5cm of k_s]				(g) {$G$};


  \edge {fi} {y} ;
  \edge {f} {fi} ;
  \edge {gp} {f} ;
  \edge {kj} {gp};
  \edge {ks} {gp};
  \edge {mu} {gp};
  \edge {zero} {mu};
  \edge {g}{k_s};
  \edge {k_s}{ks};
  \edge {theta}{ks};

  \edge {s} {gp};

  \plate {fiyi} {(y) (fi)} {$i \in 1 ... N$};
  \plate {M} {(fiyi) (f) (gp) (kj) (s)} {$j \in 1 ... M$}

\end{tikzpicture}

\caption{Graphical representation of relational ABCD. Given $M$ individual data sets, each data set contains $N$ data points which is modeled by a Gaussian process latent variable $f$. $\mathcal{GP}$ is characterized by a shared kernel $k_S$, and $M$ distinctive kernels $k_{d_j}$, and $M$ scaled factors $\sigma_j$}
\label{fig:RKL}
\end{figure}
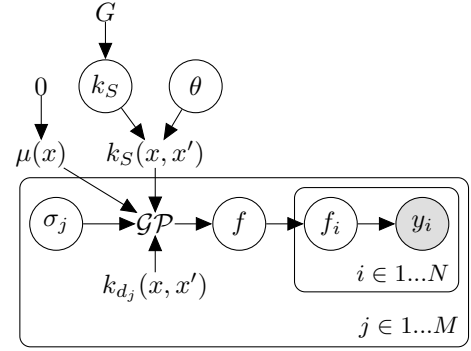

Relational ABCD considers two methods: Relational Kernel Learning, and Semi-Relational Kernel Learning (see Figure~\ref{fig:RKL}). The former is that the sequences share the same kernel structure which represents high-level properties like periodic, trends but differ by additive scale factors $b_j$ and multiplicative scale factors $v_j$. The latter relaxes the assumption among sequences by considering two parts of structure including a shared structure
and an individualized structure $k_{d_j}$. Comparing to ABCD qualitatively and quantitatively, it showed improvements in term of capturing general information across time-series as well as extrapolation performance.

%% file: 2.3-probabilistic-programming.tex
Stan \citep{stan} is similar to BUGS \citep{bugs} which enables users to write a Bayesian inference model. Stan development team provides friendly Stan's APIs for several languages (R, Python, Matlab), allowing access to many type of users. The fundamental concept used in Stan is the No U-Turn (NUTS) sampler \citep{nuts} which builds a tree of possible samples by randomly simulating Hamiltonian dynamics both forwards and backwards in time until the combined trajectory turns back on itself.

\lstset{basicstyle=\footnotesize\ttfamily,breaklines=true}
\lstset{framextopmargin=50pt,frame=bottomline}
\begin{figure}[ht!]
\begin{lstlisting}[basicstyle=\ttfamily\small]
data {
   // N >= 0
   int<lower=0> N; 
   // y[n] in { 0, 1 }
   int<lower=0,upper=1> y[N]; 
}
parameters {
   // theta in [0, 1]
   real<lower=0,upper=1> theta; 
}
model {
   // prior
   theta ~ beta(1,1); 
   // likelihood
   for (n in 1:N)
      y[n] ~ bernoulli(theta); 
}
\end{lstlisting}
\caption{A sample Stan code for estimating a Bernoulli parameter} 
\label{fig:stan}
\end{figure}

A sample Stan code is showed in Figure~\ref{fig:stan} which considers estimating the chance of success parameter for a Bernoulli distribution based on a sequence of observer binary outcomes. Here the assumption is that $N$ binary data \texttt{y[1],..., y[N]} is independent and identically distributed, with success probability \texttt{theta}. The model makes the prior assumption \texttt{beta(1,1)} on \texttt{theta}. Then, the data is fitted in the loop \texttt{for (n in 1:N)}

The followings are concentrated on how a \texttt{Stan} program is structured and executed.

\begin{itemize}
\item {\emph{Data block} Data block declares the data to fit the model. In the Figure~\ref{fig:stan}, the data block declares an integer value \texttt{N} which is the number of observed data. The array \texttt{y} has size \texttt{N}, containing information of success outcomes. It is able to set constraints of data with an upper bound \texttt{upper} and a lower bound \texttt{lower}.}
\item {\emph{Transformed data block} A transformed data block is used to define a new auxiliary variables which is computed from data, containing mediate information. Figure~\ref{fig:stan} does not include the transformed data block.\\
\emph{Parameter block} In Figure~\ref{fig:stan}, the program has only one parameter \texttt{theta} defined in the parameter block. We also can set constraints on parameters.}
\item \emph{Transformed parameter block} The transformed parameter block defines transforms of parameters for a model. It is optional, and does not appear in Figure~\ref{fig:stan}.
\item \emph{Model block} The model block defines the log probability function on the parameter space. In the program, by providing the prior on \texttt{theta}, the likelihood is evaluated.
\item \emph{Generated quantities block} The (optional) generated quantities allows values that depend on parameters and data. It may be used to calculate predictive inferences. It can carry out forward simulation for predictive posterior checks.
\end{itemize}

%% file: 3-combi.tex
\par
\begin{figure*}[!ht]
\centering
\includegraphics[width=0.65\textwidth]{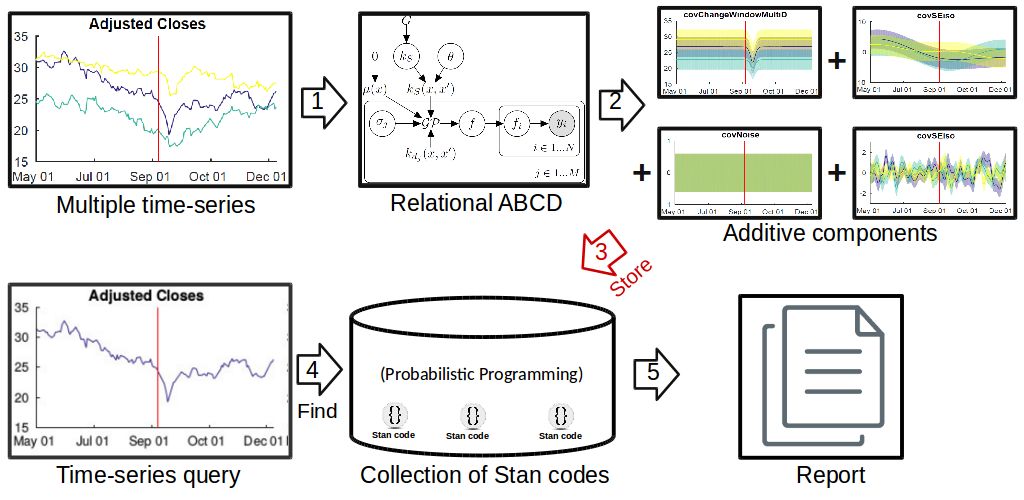}
\caption{An overview of the system in which the relational ABCD framework (or ABCD framework) combines with probabilistic programming. Step 1 executes the input in the relational ABCD framework (ABCD framework); Step 2 retrieves the output as a kernel; Step 3 generates probabilistic programs; Step 4 executes a query into system; Step 5 automatically makes a report with respect to query. Step 1, 2 are procedures in the relational ABCD framework (ABCD framework). We address the step 3 in this paper.}
\label{fig:model}
\end{figure*}

 Generating probabilistic programming from ABCD and/or relational ABCD is a crucial component in the system we aim to build (see Figure \ref{fig:model}).  Prior to this component, one can choose either ABCD or relational ABCD based on whether the preference is for a single time series or for global information in multiple time series. Both ABCD and relational ABCD play as producers which output compositional kernels from data (Step 1 and 2). Step 4 and 5 are an example application. We will discuss what is inside Step 3 in this section.
\par From now on, we use the notation \texttt{StanABCD} to indicate the \texttt{Stan} probabilistic programs generated automatically from ABCD.

\subsection{Base kernels}
An ABCD's result is represented by a compositional structure which is a sum of products of base kernels. This summation is the outcome of simplifications: the multiplication of two SE kernel produces another SE with different parameter values. The product of WN and any stationary kernel including C, PER, WN, SE results a new WN kernel. Multiplying C with any kernel does not change the kernel but changes the scale parameter of that kernel \citep{lloyd2014automatic}. Hence, let $G$ be a set of all possible kernel expressions, written as
$$G = \{\sum k \prod_m \textrm{LIN}^{(m)} \sigma^{(n)}\}$$
where $\sigma$ is the sigmoid function, $k$ is in 
$$K = \{\textrm{WN}, \textrm{C}, \prod_k\textrm{PER}^{(k)}, \textrm{SE}\prod_k\textrm{PER}^{(k)}\}$$

For example, a learned compositional kernel is described as
\begin{equation}
\label{eq:2}
\textrm{SE} \times \textrm{LIN}
\end{equation}
Here, the square exponential kernel and linear kernel are written respectively as 
$$\textrm{SE}(x, x') = \sigma^2 \exp\left( - \frac{(x - x')^2}{2l^2} \right),$$
$$\textrm{LIN}(x, x') = \sigma^2(x - l)(x' - l).$$
The kernel (\ref{eq:2}) can be understood linguistically as 'a smooth function with linearly (LIN) increasing amplitude'.\\

Another real-world example is a chosen currency exchange data set (see Figure \ref{fig:indo}). The data set contains exchange value of Indonesian Rupiad from 2015-06-28 to 2015-12-30 acquired from Yahoo Finance \citep{yahoofinance}. Carried experiments on this data set, relational ABCD found the best compositional kernel which is shortly written as 
\begin{equation}
\textrm{CW}(\textrm{SE} + \textrm{CW}(\textrm{WN} + \textrm{SE},\textrm{WN} ), \textrm{C} )
\label{eq:indo}
\end{equation}
This kernel is well-explained for several currency exchanges sharing common financial behaviors. A qualitative result shows that the changewindow (CW) kernel occurs around mid September 2015 which reflects big financial events (FED’s announcement about policy changes in interest rates, China’s foreign exchange reserves falls) \citep{1511.08343} . We take this compostional kernel as a typical example for demonstration purpose.

Given a data set and a learned kernel, we are interested in encoding them into a \texttt{Stan} program. In order to do that, we first prepare built-in base kernels in \texttt{Stan} version. A base kernel is written as a \texttt{Stan} function which takes data, and hyperparameters as input and returns a matrix. The matrix has elements reflecting how similar (correlated) the data points in data set are. Each base kernel have a specific number of hyperparameters itself. For the SE kernel case, it has two hyperparameters: a scale factor $\sigma$, and a lengthscale $l$; then we build a \texttt{Stan} code as showed in Figure~\ref{fig:se} (The detail implementations of other kernels are in the Appendix).
\begin{figure}[!ht]
\input{./kernel/se.tex}
\caption{The implementation of SE kernel on \texttt{Stan}}
\label{fig:se}
\end{figure}
\input{./code/kernel.tex}
Next, we will discuss how a \texttt{StanABCD} organize. \texttt{StanABCD}s share common conventional blocks (as in previous section) but the compositional kernel. We briefly describe what is required in each blocks.

\subsubsection{Data block}
\begin{figure}[h!]
\includegraphics[width=0.5\textwidth]{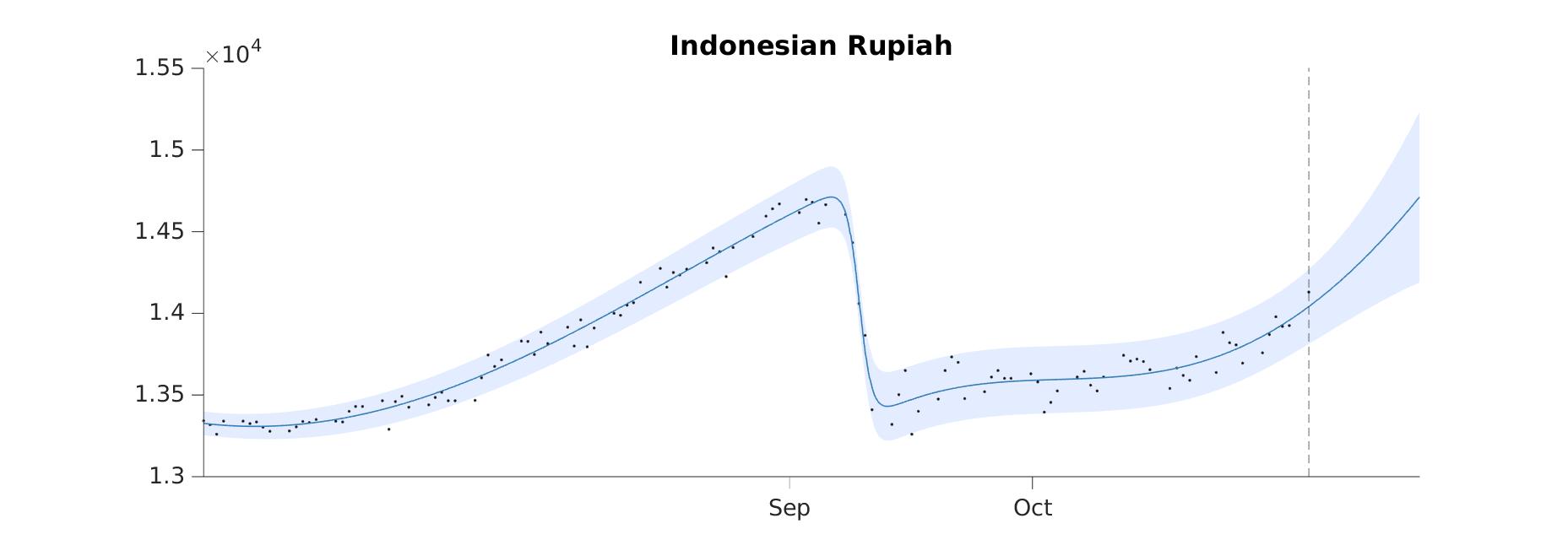}
\caption{Indonesian Rupiad exchange data. Dot: raw data point. Line with shade: Gaussian Process prediction with 95\% confidence region. Vertical dash line separates the training data and test data.}
\label{fig:indo}
\end{figure}

In general, \texttt{StanABCD} contains training data points $X$ and test data points $X_{\star}$. Abiding by the \texttt{Stan} convention, \texttt{StanABCD} declares data as following: 
\input{./code/data_block.tex}
Here, we provide the information of training data through \texttt{N1} (number of training data points) and vectors \texttt{x1}, \texttt{y1}. Similarly, test data is specified by the number of test data points \texttt{N2} and a vector \texttt{x2}. 
For example, we analyze Indonesian Rupiah exchange data with the period from July 2015 to December 2015 as shown in Figure \ref{fig:indo}. This data set consists of 132 data points in which we take the first 120 data points as training data, and the next 12 data points as test data. We have \texttt{N1} = 120, \texttt{x1} be the days in training data set, \texttt{y1} be the exchange value, \texttt{N2} = 12, and \texttt{x2} be the days in test data set. We want to predict the exchange value on \texttt{x2}.\\

\subsubsection{Parameter block and transformed parameters block}
These blocks consist of all necessary parameters to construct the model. With the purpose of sampling data on test points,  
\texttt{StanABCD} has the parameter block containing a parameter \texttt{z} as an array with length \texttt{N2} equal to the number of test data points. \texttt{z} follows unit normal distribution to increase the sampling performance which is discussed later in the transformed parameters block. If we use \texttt{StanABCD} not only as a sampler, the parameter block should be customized by adding more parameters for our desired models.

What the transformed parameters block does is to make computation for the posterior distribution of GPs. Based on GP prior, the joint distribution of training output $\mathbf{y}$ and test output $\mathbf{y}_\star$ is represented as
\begin{equation*}
\begin{pmatrix}
\mathbf{y}\\
\textbf{y}_\star
\end{pmatrix}
\sim
\mathcal{N}\left(
\mathbf{0},
\begin{bmatrix}
K(X,X) & K(X,X_\star) \\
K(X_\star, X) & K(X_\star, X_\star)
\end{bmatrix}
\right).
\end{equation*}
Assume we already know the structure of compositional kernel $K(.,.)$ from ABCD. 
From this compositional kernel structure, $K(X,X_\star)$ is a $n \times n_\star$ matrix evaluated at all pairs of training and test points, where $n = |X|$ is the number of training data points, $n_\star = |X_\star|$  is the number of test data points. Analogously, we compute $K(X,X)$, $K(X_\star, X)$, and $K(X_\star,X_\star)$. Note that $K(X,X_\star) = K(X_\star, X)^T$. Now, using the conditioning Gaussians, it follows that
\begin{equation}
\begin{split}
\mathbf{y}_\star | X_\star, X, \mathbf{y} & \sim \mathcal{N}(K(X, X_\star)K(X, X)^{-1}\mathbf{y}, \\
& K(X_\star, X_\star) - K(X, X_\star)K(X,X)^{-1}K(X,X_\star))
\end{split}
\label{eq:post}
\end{equation}

The posterior distribution is analytically tractable. This makes easier to represent on a \texttt{Stan} programming because it supports most of distributions. Here, we only need to specify a mean $\mu = K(X, X_\star)K(X, X)^{-1}\mathbf{y}$ and a covariance matrix $\Sigma = K(X_\star, X_\star) - K(X, X_\star)K(X,X)^{-1}K(X,X_\star)$, in order to declare a multivariate normal distribution in the next blocks.  

Taking a consideration about the efficiency of implementation,  the Cholesky decomposition (which is available as a built-in function in \texttt{Stan}) of $\Sigma$ is pre-computed. Let us denote the Cholesky decompositon of $\Sigma$ be $L$. \texttt{Stan} only perform its sampling method to produce a unit normal distribution
\begin{equation}
z \sim \mathcal{N}(\mathbf{0}, \mathbf{I})
\label{eq:unit_normal}
\end{equation}

Then, we transform $z$ into $\mu + Lz$ to obtain our target distribution
$$\mu + Lz \sim \mathcal{N}(\mu, \Sigma = LL^T).$$

Here is the sample code for this block:
\input{./code/trans.tex}
The above variables \texttt{Sigma}, \texttt{Omega}, \texttt{K} correspond respectively to the terms $K(X,X)$, $K(X_\star, X_\star), K(X, X_\star)$. 

We still leave the question how to build a compositional kernel from base kernels. Basically, the compositional kernel is the sum of product of base kernels. To represent the compositional kernel in \texttt{Stan}, the summation is used as a matrix sum operation and the multiplication between base kernels is replaced by the Hadamard product. Here is an example

\input{./code/kernel.tex}
\subsubsection{Model block}
\texttt{Stan} allows us to quickly design a Bayesian hierarchical model. Utilizing the mean and covariance computed in the previous  block helps us declare a normal distribution which plays a role as the first level in the multiple levels of hierarchical model. However, we set this aside and only illustrate the case that we sample the posterior distribution on test data points. From (\ref{eq:unit_normal}), we declare a  Gaussian distribution $\mathcal{N}(\mathbf{0}, \mathbf{I})$ to serve the sampling purpose in the generated quantities block. 
\input{./code/model.tex}
\subsubsection{Generated quantities block} 
For purpose of generating sample extrapolation value, the normal distribution declared in the model block will be called one time. 
\input{./code/generate.tex}
In order to get the sample of test output \texttt{y2}, the above \texttt{Stan} code performs the linear transformation on sample values generated from \texttt{z} ($\mathcal{N}(\mathbf{0}, \mathbf{I})$) as we explained in the transformed parameters block.

%% file: code/kernel.tex
\lstset{basicstyle=\footnotesize\ttfamily}
\lstset{framextopmargin=50pt}
\begin{lstlisting}    
matrix KERNEL(vector x, vector y){
   return LIN(x, y, 0.39, 1945.15) .*(CONST(x, y, 3.10) + PER(x, y, 0.52, 1.00, 0.00) .*(WN(x, y, 5.53) + SE(x, y, 297.83, 517.18))) .*(CONST(x, y, 595.15) + SE(x, y, 2.53, 0.56));
}
\end{lstlisting}

%% file: code/data_block.tex
\lstset{basicstyle=\footnotesize\ttfamily}
\lstset{framextopmargin=50pt}
\begin{lstlisting}
data {
   int<lower = 1> N1;
   vector[N1] x1;
   vector[N1] y1;
   int<lower=1> N2;
   vector[N2] x2;
}
\end{lstlisting}

%% file: code/trans.tex
\lstset{basicstyle=\footnotesize\ttfamily}
\lstset{framextopmargin=50pt}
\begin{lstlisting}
transformed parameters {
   vector[N2] mu;
   matrix[N2,N2] L;
   {
     matrix[N1, N1] Sigma;
     matrix[N2, N2] Omega;
     matrix[N1, N2] K;
     matrix[N2, N1] K_transpose_div_Sigma;
     matrix[N2, N2] Tau;
     Sigma <- KERNEL(x1,x1);
     Omega <- KERNEL(x2,x2);
     K <- KERNEL(x1,x2);
     K_transpose_div_Sigma <- K' / Sigma;
     mu <- K_transpose_div_Sigma * y1;
     Tau <- Omega - K_transpose_div_Sigma*K;
     for (i in 1:N2)
        for(j in (i + 1):N2)
          Tau[i,j] <- Tau[j, i];
     L <- cholesky_decompose(Tau);
   }
}
\end{lstlisting}

%% file: code/model.tex
\begin{lstlisting}
   z ~ normal(0,1);
\end{lstlisting}

%% file: code/generate.tex
\begin{lstlisting}
generated quantities {
   vector[N2] y2;
   y2 <- mu + L * z;
}
\end{lstlisting}

%% file: 4-experiment.tex
\subsubsection{Data set} Beside the Indonesian Rupiah exchange data mentioned in previous section, we select a airline data set to perform experiments on. The data set describes monthly international airline passenger numbers for the period between January 1949 and December 1960 (\cite{Box}). The number of passengers was periodic with a typical period 1 year. The total number of passengers per year increased monotonically. ABCD captures this information well, and explain the data set by a compositional kernel: LIN + SE $\times$ PER $\times$ LIN + SE + WN $\times$ LIN.

\begin{figure}[!t]
    \centering
    \begin{subfigure}[b]{0.3\textwidth}
        \includegraphics[width=\textwidth]{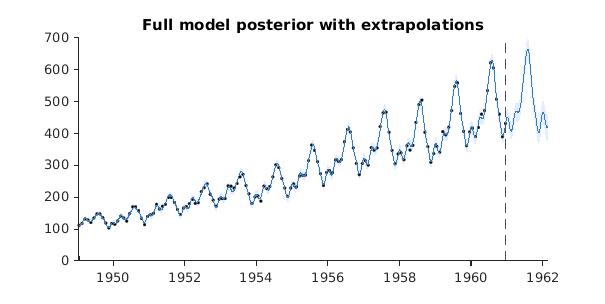}
        \caption{}
    \end{subfigure}
    ~ 
    \begin{subfigure}[b]{0.3\textwidth}
        \includegraphics[width=\textwidth]{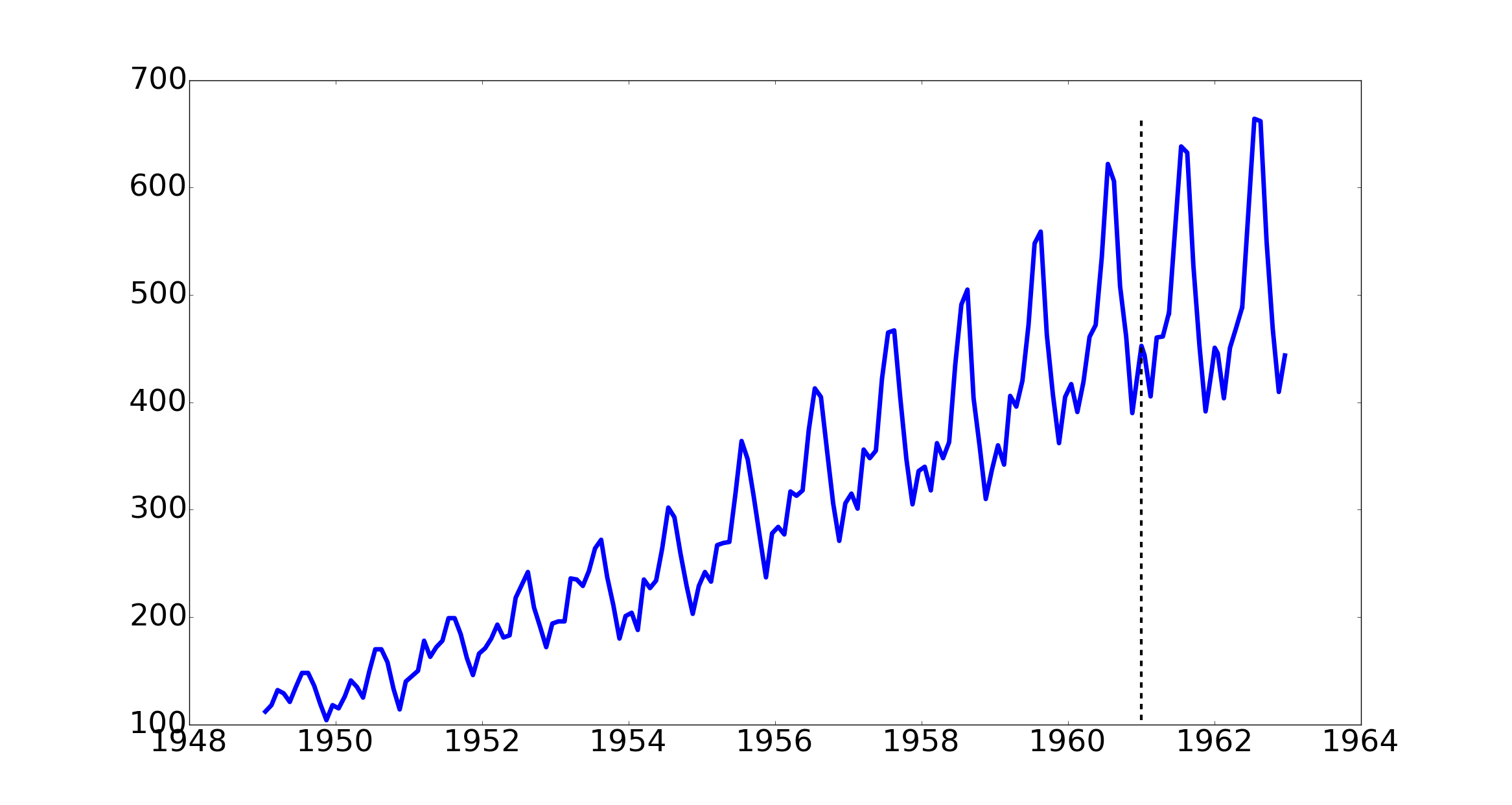}
        \caption{}
    \end{subfigure}
    ~ 
    \caption{A comparison of ABCD result and \texttt{StanABCD}'s sample extrapolation (from dash line). a) Extraplation of airline data from ABCD \citep{lloyd2014automatic}. b) Generated sample from \texttt{StanABCD}}
    \label{fig:exp}
\end{figure}
\subsubsection{Sampling data}
\begin{figure}
\centering
\includegraphics[width=0.3\textwidth]{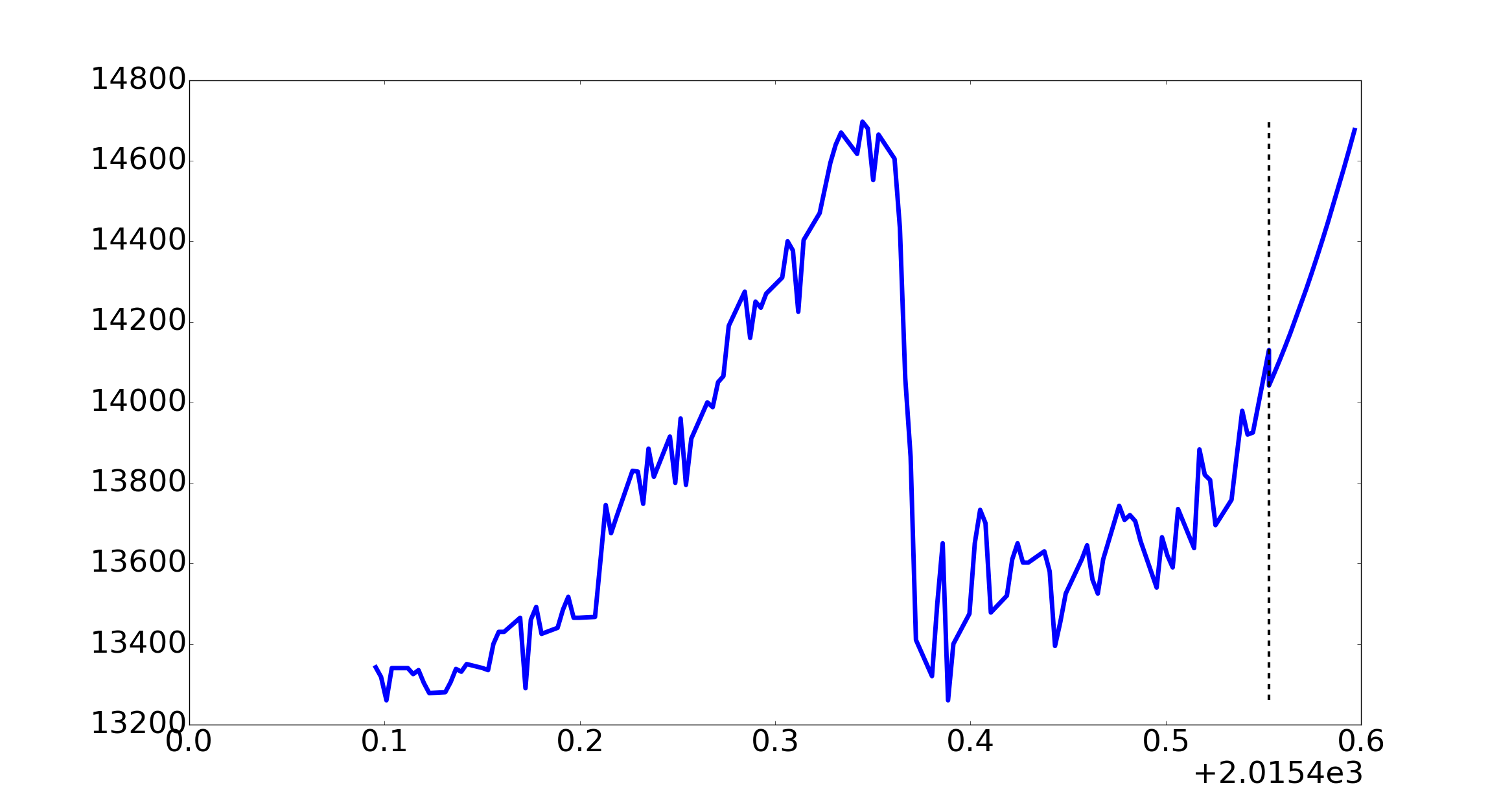}
\caption{The extrapolation sampling (from vertical dash line) from \texttt{StanABCD} for Indonesian Rupiah data set. Comparing to ABCD result (see Figure \ref{fig:indo}), it can achieve a similar result.}
\label{fig:exp_indo}
\end{figure}

We provide a complete sample \texttt{Stan} code in the appendix. We want to get extrapolation sample values on test points (from January 1961) of airline data set. During the experiment, we use Python to retrieve data set then pass to \texttt{Stan} compiler through \textbf{PyStan} \citep{pystan}. Figure \ref{fig:exp} and \ref{fig:exp_indo} show that \texttt{StanABCD} provides similar results as ABCD in the view of extrapolation performance. Our generating method guarantees a reliable way to perform one-to-one mapping from ABCD result into a probabilistic programming.

%% file: 5-conclusion.tex
We propose a beautiful blend between the automatic statistician and probabilistic programming. As the result, it opens a broad direction to explore on encoded \texttt{Stan} programs because of their potentiality to make further inference or perform statistical relational learning. On the other hand, \texttt{StanABCD} provides a promising way to accelerate the learning kernel in ABCD framework which requires an exhaustive search procedure. A database of \texttt{StanABCD}s is one of the possible solutions. 

%% file: 6-appendix.tex
\appendix
\section*{Appendix}
\subsection{Base kernels}
\subsubsection{White noise kernel}
A white noise kernel is written as
$$\textrm{WN}(x, x') = \sigma^2 \delta_{x,x'}$$
We implement this kernel in 
\input{./kernel/wn.tex}
\subsubsection{Constant kernel}
A constant kernel is written as
$$\textrm{C}(x, x') = \sigma^2$$
\input{./kernel/const.tex}
\subsubsection{Linear kernel}
A linear kernel is written as
$$\textrm{LIN}(x, x') = \sigma^2(x - l)(x' - l)$$
\input{./kernel/lin.tex}
\subsubsection{Squared exponential kernel}
A squared exponential is defined as follows:
$$\textrm{SE}(x, x') = \sigma^2 \exp\left( - \frac{(x - x')^2}{2l^2} \right)$$
\input{./kernel/se.tex}
\subsubsection{Periodic kernel}
A periodic kernel is defined as 
$$\textrm{PER}(x,x') = \sigma^2 \frac{\exp \left(\frac{\cos \frac{2\pi(x - x')}{p}}{l^2}\right) - I_0(\frac{1}{l^2})}{\exp(\frac{1}{l^2}) - I_0(\frac{1}{l^2})}, $$
where $I_0$ is the modified Bessel function of the first kind of order zero
\input{./kernel/per.tex}
\subsubsection{Changepoint operator}
A changepoint operator on kernels $k_1$ and $k_2$ is defined as 
\begin{equation*}
\begin{split}
\textrm{CP}(k_1,k_2)(x,x') = & \sigma(x)k_1(x,x')\sigma(x') + \\
& (1- \sigma(x))k_2(x,x')(1 - \sigma(x'))
\end{split}
\end{equation*}
where $\sigma(x) = \frac{1}{2}(1 + \textrm{tanh}(\frac{l - x}{s}))$
\input{./kernel/changepoint.tex}
\newpage
\onecolumn
\subsection{A sample code}
\input{./code/complete.tex}

%% file: kernel/wn.tex
\lstset{basicstyle=\footnotesize\ttfamily,breaklines=true}
\lstset{framextopmargin=50pt,frame=bottomline}
\begin{lstlisting}
matrix WN(vector x, vector y, real scale){
   matrix[rows(x), rows(y)] m;
   for (i in 1: rows(x))
      for (j in 1: rows(y)){
         m[i,j] <- if_else(i == j, scale, 0.0);
      }
   return m;
}
\end{lstlisting}

%% file: kernel/const.tex
\lstset{basicstyle=\footnotesize\ttfamily,breaklines=true}
\lstset{framextopmargin=50pt,frame=bottomline}
\begin{lstlisting}
matrix CONST(vector x, vector y, real constant){
   matrix[rows(x), rows(y)] m;
   for (i in 1: rows(x))
      for (j in 1: rows(y)){
         m[i,j] <- constant;
      }
   return m;
}
\end{lstlisting}

%% file: kernel/lin.tex
\lstset{basicstyle=\footnotesize\ttfamily,breaklines=true}
\lstset{framextopmargin=50pt,frame=bottomline}
\begin{lstlisting}
matrix LIN(vector x, vector y, real sf, real location){
   matrix[rows(x), rows(y)] m;
   for (i in 1: rows(x))
      for (j in 1: rows(y)){
         m[i,j] <- sf *(x[i] - location) * (y[j] - location);
      }
   return m;
}
\end{lstlisting}

%% file: kernel/se.tex
\lstset{basicstyle=\footnotesize\ttfamily,breaklines=true}
\lstset{framextopmargin=50pt,frame=bottomline}
\begin{lstlisting}
matrix SE(vector x, vector y, real sf, real l){
   matrix[rows(x), rows(y)] m;
   for (i in 1: rows(x))
      for (j in 1: rows(y)){
         m[i,j] <- sf * exp(-pow(x[i] - y[j], 2) / (2*l) );
      }
   return m;
}
\end{lstlisting}

%% file: kernel/per.tex
\lstset{basicstyle=\footnotesize\ttfamily,breaklines=true}
\lstset{framextopmargin=50pt,frame=bottomline}
\begin{lstlisting}
real covD(real d, real ell2){
   real c;
   real temp;
   real b0;
   if (sqrt(ell2) > 10000){
      c <- cos(d);
   }else if (1.0 /ell2 < 3.75){
      temp <- exp(cos(d)/ell2);
      b0 <- bessel_first_kind(0, 1/ell2);
      c <- (temp - b0) /(exp(1/ell2) - b0);
   }else {
      temp <- exp((cos(d) - 1)/ell2);
      b0 <- embi0(1/ell2);
      c <- (temp - b0)/(1 - b0);
   }
   return c;
}
matrix PER(vector x, vector y, real lengthscale, real period, real sf){
   real d;
   matrix[rows(x), rows(y)] m;
   for (i in 1: rows(x))
      for (j in 1: rows(y)){
         d <- 2 * pi()*(x[i] - y[j])/period;
         m[i,j] <- sf * covD(d, lengthscale);
      }
   return m;
}
\end{lstlisting}

%% file: kernel/changepoint.tex
\lstset{basicstyle=\footnotesize\ttfamily,breaklines=true}
\lstset{framextopmargin=50pt,frame=bottomline}
\begin{lstlisting}
matrix CP(vector x, vector y, real location, real steepness, matrix kernel1, matrix kernel2){
   matrix[rows(x), rows(y)] m;
   matrix[rows(x), 1] sigmoid_x;
   matrix[rows(y), 1] sigmoid_y;
   sigmoid_x <- sigmoid(x, location, steepness);
   sigmoid_y <- sigmoid(y, location, steepness);
   for (i in 1: rows(x)){
      for(j in 1: rows(y)){
         m[i,j] <- sigmoid_x[i, 1]*kernel1[i,j]*sigmoid_y[j,1] + (1 - sigmoid_x[i,1])*kernel2[i,j]*(1-sigmoid_y[j,1]);
      }
   }
   return m;
}
\end{lstlisting}

%% file: code/complete.tex
\lstset{basicstyle=\footnotesize\ttfamily}
\lstset{framextopmargin=50pt}
\begin{lstlisting}
functions {
    matrix CONST(vector x, vector y, real constant){

        matrix[rows(x), rows(y)] m;
        for (i in 1: rows(x))
            for (j in 1: rows(y)){
                m[i,j] <- constant;
            }
        return m;
    }

    matrix LIN(vector x, vector y, real sf, real location){

        matrix[rows(x), rows(y)] m;
        for (i in 1: rows(x))
            for (j in 1: rows(y)){
                m[i,j] <- sf *(x[i] - location) * (y[j] - location);
            }

        return m;
    }

    real embi0(real x){ #= exp(-x)*besseli(0,x) => 9.8.2 Abramowitz & Stegun
        real y;
        y <- 3.75/x;
        y <- 0.39894228     + 0.01328592*y   + 0.00225319*y^2 - 0.00157565*y^3
            + 0.00916281*y^4 - 0.02057706*y^5 + 0.02635537*y^6 - 0.01647633*y^7
            + 0.00392377*y^8;
        y <- y/sqrt(x);

        return y;
    }

    real covD(real d, real ell2){
        real c;
        real temp;
        real b0;

        if (sqrt(ell2) > 10000){
            c <- cos(d);
        }else if (1.0 /ell2 < 3.75){
            temp <- exp(cos(d)/ell2);
            b0 <- bessel_first_kind(0, 1/ell2);
            c <- (temp - b0) /(exp(1/ell2) - b0);
        }else {
            temp <- exp((cos(d) - 1)/ell2);
            b0 <- embi0(1/ell2);
            c <- (temp - b0)/(1 - b0);
        }
        return c;
    }

    matrix PER(vector x, vector y, real lengthscale, real period, real sf){
        real d;

        matrix[rows(x), rows(y)] m;
        for (i in 1: rows(x))
            for (j in 1: rows(y)){
                d <- 2 * pi()*(x[i] - y[j])/period;
                m[i,j] <- sf * covD(d, lengthscale);
            }
        return m;
    }

    matrix SE(vector x, vector y, real sf, real lengthscale){

        matrix[rows(x), rows(y)] m;
        for (i in 1: rows(x))
            for (j in 1: rows(y)){
                m[i,j] <- sf * exp(-pow(x[i] - y[j], 2) / (2*lengthscale) );
            }

        return m;
    }

    matrix WN(vector x, vector y, real scale){

        matrix[rows(x), rows(y)] m;

        for (i in 1: rows(x))
            for (j in 1: rows(y)){
                m[i,j] <- if_else(i == j, scale, 0.0);
            }

        return m;
    }

    matrix sigmoid(vector x, real l, real s) {
        matrix[rows(x), 1] sig;
        for (i in 1: rows(x)){
            sig[i, 1] <- 0.5 *(1.0 + tanh((l - x[i])*s)); #reference to matlab code
        }
        return sig;
    }

     matrix sigmoid_cw(vector x, real l, real s, real w) {
        matrix[rows(x), 1] sig;
        for (i in 1: rows(x)){
            sig[i, 1] <- 0.25 *(1.0 + tanh((x[i] - (l - 0.5*w))*s))*(1.0 + tanh(-(x[i] - (l + 0.5*w))*s)); #reference to matlab code
        }
        return sig;
    }

    matrix ones(int rows){
        matrix[rows, 1] m;
        for(i in 1:rows)
            m[i, 1] <- 1.0;

        return m;
    }

    matrix CP(vector x, vector y, real location, real steepness, matrix kernel1, matrix kernel2){

        matrix[rows(x), rows(y)] m;
        matrix[rows(x), 1] sigmoid_x;
        matrix[rows(y), 1] sigmoid_y;
        sigmoid_x <- sigmoid(x, location, steepness);
        sigmoid_y <- sigmoid(y, location, steepness);

        for (i in 1: rows(x)){
            for(j in 1: rows(y)){
                m[i,j] <- sigmoid_x[i, 1]*kernel1[i,j]*sigmoid_y[j,1] + (1 - sigmoid_x[i,1])*kernel2[i,j]*(1-sigmoid_y[j,1]);
            }
        }
        return m;
    }

    matrix CW(vector x, vector y, real location, real steepness, real width, matrix kernel1, matrix kernel2){

        matrix[rows(x), rows(y)] m;
        matrix[rows(x), 1] sigmoid_x;
        matrix[rows(y), 1] sigmoid_y;
        sigmoid_x <- sigmoid_cw(x, location,steepness, width);
        sigmoid_y <- sigmoid_cw(y, location,steepness, width);

        for (i in 1: rows(x)){
            for(j in 1: rows(y)){
                m[i,j] <- sigmoid_x[i, 1]*kernel1[i,j]*sigmoid_y[j,1] + (1 - sigmoid_x[i,1])*kernel2[i,j]*(1-sigmoid_y[j,1]);
                #print(i, " ", sigmoid_x[i, 1], "---", j, " ", sigmoid_y[j,1]);
            }
        }

        #m <- (sigmoid_x* sigmoid_y') .* kernel1 + ((ones(rows(x)) - sigmoid_x)*(ones(rows(y))-sigmoid_y)').*kernel2;

        return m;
    }

    matrix RQ(vector x, vector y, real sf, real lengthscale, real alpha){

        matrix[rows(x), rows(y)] m;
        for (i in 1: rows(x))
            for (j in 1: rows(y)){
                m[i,j] <- sf * pow(1 + pow(x[i] - y[j],2)/(2*lengthscale*alpha), -alpha);
            }

        return m;
    }

    matrix KERNEL(vector x, vector y){
		return LIN(x, y, 0.394302019493, 1945.14751497) .*(CONST(x, y, 3.09970709195) + PER(x, y, 0.518041024808, 1.00227799156, 0.000288003211837) .*(WN(x, y, 5.53246245772) + SE(x, y, 297.831942207, 517.179707302))) .*(CONST(x, y, 595.14997953) + SE(x, y, 2.53149893027, 0.555234284261));
            }
        }
        
data {
    int<lower = 1> N1;
    vector[N1] x1;
    vector[N1] y1;
    int<lower=1> N2;
    vector[N2] x2;

}

parameters {
    vector[N2] z;
}

transformed parameters {

    vector[N2] mu;
    matrix[N2,N2] L;

    {
        matrix[N1, N1] Sigma;
        matrix[N2, N2] Omega;
        matrix[N1, N2] K;

        matrix[N2, N1] K_transpose_div_Sigma;
        matrix[N2, N2] Tau;

        Sigma <- KERNEL(x1,x1);
        Omega <- KERNEL(x2,x2);
        K <- KERNEL(x1,x2);

        K_transpose_div_Sigma <- K' / Sigma;
        mu <- K_transpose_div_Sigma * y1;
        Tau <- Omega - K_transpose_div_Sigma*K;

        for (i in 1:N2)
            for(j in (i + 1):N2)
                Tau[i,j] <- Tau[j, i];

        L <- cholesky_decompose(Tau);
    }
}

model{
     z ~ normal(0,1);
}

generated quantities {
    vector[N2] y2;
    y2 <- mu + L * z;
}
\end{lstlisting}